\documentclass[10pt,twocolumn,letterpaper]{article}

\usepackage{cvpr}
\usepackage{times}
\usepackage{epsfig}
\usepackage{graphicx}
\usepackage{amsmath}
\usepackage{amssymb}
\usepackage{listings}
\usepackage{float}

\usepackage[breaklinks=true,bookmarks=false]{hyperref}

\cvprfinalcopy % *** Uncomment this line for the final submission

\def\cvprPaperID{****} % *** Enter the CVPR Paper ID here

\begin{document}

%%%%%%%%% TITLE
\title{A Free Lunch in Generating Datasets:\\Building a VQG and VQA System with Attention and Humans in the Loop}

\author{Sho Arora, Jihyeon Lee\\
Stanford University\\
{\tt\small shoarora, jihyeon @ cs.stanford.edu}
% \and
}

\maketitle
%\thispagestyle{empty}

%%%%%%%%% ABSTRACT
\begin{abstract}
	Despite their importance in training artificial intelligence systems, large datasets remain challenging to acquire. For example, the ImageNet dataset~\cite{deng2009imagenet} required fourteen million labels of basic human knowledge, such as whether an image contains a chair. Unfortunately, this knowledge is so simple that it is tedious for human annotators but also tacit enough such that they are necessary. However, human collaborative efforts for tasks like labeling massive amounts of data are costly, inconsistent, and prone to failure, and this method does not resolve the issue of the  resulting dataset being static in nature. What if we asked people questions they \textit{want} to answer and collected their responses as data? This would mean we could gather data at a much lower cost, and expanding a dataset would simply become a matter of asking more questions. We focus on the task of Visual Question Answering (VQA) and propose a system that uses Visual Question Generation (VQG) to produce questions, asks them to social media users, and collects their responses. We present two models that can then parse clean answers from the noisy human responses significantly better than our baselines, with the goal of eventually incorporating the answers into a Visual Question Answering (VQA) dataset. By demonstrating how our system can collect large amounts of data at little to no cost, we envision similar systems being used to improve performance on other tasks in the future. 
\end{abstract}

%%%%%%%%% BODY TEXT
\section{Introduction}
Modern artificial intelligence systems rely on massive amounts of human-labeled training data, but generating these datasets remains a challenge. For example, the ImageNet dataset~\cite{deng2009imagenet} required fourteen million labels of basic human knowledge, such as whether an image contains a chair. Unfortunately, this knowledge is both so simple that it is tedious for humans to label but also tacit enough such that human annotators are necessary. This combination of tedious and tacit makes labels difficult to acquire, and the current method of organizing human collaborative efforts is not sustainable. Not only are such efforts costly, results are inconsistent and prone to failure; they also still do not solve the issue of the final output being limited in scope and static ~\cite{healy2003ecology,hill2013almost,reich2012state}.

What if, instead of organizing people specifically to label massive amounts of data, people are asked questions that they \textit{want} to answer and their responses are collected as data? This would enable us to gather data at a significantly lower cost and expand datasets more easily, by asking more questions. To test this idea, we focus on the task of Visual Question Answering (VQA). The task of VQG is to generate questions based on image content, while VQA is to comprehend and answer questions based on image content \cite{DBLP:journals/corr/ZhouTSSF15,DBLP:journals/corr/AntolALMBZP15,DBLP:journals/corr/MostafazadehMDZ16,DBLP:journals/corr/ZhangQYYZ16}. Our goal is to create a model is able to discover visual concepts with which it is currently unfamiliar by asking questions to social media users about their photos and expanding its knowledge of those concepts.  We propose a system that uses Visual Question Generation (VQG) to produce questions, asks them to social media users, and retrieves the noisy responses to clean them to be eventually incorporated into a VQA dataset. 

For this class project, we focus our evaluation on how well we can parse clean answers from noisy human responses because they are the information that will eventually be used to expand our dataset. We collect ground truth clean answers from Amazon Mechanical Turk workers (more details provided in the later Data section), and the metric we want to optimize is the BLEU score between the ground truth and the answers our model has produced \cite{bleu}. 

To summarize our results, we propose two different models that were able to significantly improve BLEU scores compared to the baselines. The first model clusters responses and then extracts latent templates based on the most frequently occurring bigrams/trigrams, and it obtained a 40x improvement. The second model is a Recurrent Neural Network (RNN) that takes a sequence input, the noisy response, and predicts question type, which we glean by clustering the questions. We provide more motivation for why we predict question type in the Methods section, but essentially this model enabled us to overcome the challenge of having not enough data for a fully supervised approach and instead perform unsupervised learning. This deep model achieved an improvement of 50x compared to the baselines.

By leveraging VQG to generate questions that engage people and get information about unfamiliar concepts, we propose a model that collects people's responses and parses useful information from them to expand datasets for the task of VQA. We envision that this kind of system will enable AI systems to gather training data at a much lower cost.
%------------------------------------------------------------------------
\section{Related Works}

 We discuss three categories of related work: systems that targeted the same goal of creating non-static datasets, intelligent agents that interacted with humans, and the state of VQA.
 
 \subsection{Expanding Datasets}
 There have been previous attempts to create never-ending datasets. First, Never Ending Language Leaner (NELL) was proposed as a method to continuously extract information from unstructured text on the web and add to a knowledge base of structured facts \cite{nell}. The method is very similar to bootstrapping, in that the input is an ontology with pre-specified information, like contextual patterns (e.g. "cities such as "), and incrementally expands the set of trusted patterns for categories of interest. A recent case study showed how NELL has expanded the noun phrases and predicates for 293 categories specified by the initial ontology with certain confidence levels \cite{nell_case}. One shortcoming we see with NELL is that the "never-ending" is assumed from how the internet is constantly growing or how the system learns new patterns. NELL does not address the fact that much of the information needed to amass datasets is tacit; the internet is laden with assumptions that humans have, but not machines. Although it does obtain new patterns, NELL needs predefined categories as input, meaning it cannot discover entirely new concepts on its own. It also does not have a goal for its learning; for example, in our case, our goal is to produce a dataset that improves performance on the task of VQA.
 
 While NELL focused on language, Never Ending Image Learner (NEIL) proposed a semi-supervised method of extracting labeled examples of object categories with bounding boxes, scenes, and attributes \cite{neil}. NEIL employs a similar method of bootstrapping, starting with seed images for a given category, training classifiers for each subcategory, and collecting the most confidently labeled images to add to the dataset and retrain the classifiers \cite{neil}. Because NEIL leveraged the ontology of NELL to compile its categories, it has the similar shortcoming of being unable to discover new concepts easily, especially if there is not sufficient seed data. We believe our system overcomes this by asking about images that are already exist but are uncertain to the model. We also see a difference in that NEIL achieves breadth for the task of object detection, while we narrow in on collecting data for the task of VQA.
 
 The authors of \cite{noisywebvideos} presented a method to collect labeled data for human action classes from noisy web data. They use a reinforcement learning-based approach to select examples for training a classifier from web search results and learn a data labeling policy \cite{noisywebvideos}. The policy is then used to automatically label concepts for new visual concepts \cite{noisywebvideos}. Although this part of our vision is outside the scope of this class project, we hope to use a similar approach of using reinforcement learning to learn what questions are considered engaging. For example, our system could launch questions and calculate rewards based on whether or not people respond and/or provide good answers. We share their goal of scaling to the full long-tailed distribution of classes rather than relying on a set of predefined categories, although we leverage humans instead of web search to get our labels.
 
 \subsection{Conversational Agents}
 Since our system interacts with people, we also examine a few conversational agents. In 2016, Microsoft launched a chat bot, Tay, which learned to post inflammatory and offensive tweets \cite{taynews}. In a case study, \cite{taystudy} found that the intentions of designers for how Tay would be used were completely different from the expectations of Twitter users. One important takeaway is that users will unconsciously apply social rules to technologies and interact with them as if they are living entities \cite{taystudy}. Xiaoice, another chat bot, engaged over 20 million users in China and was notably good at covering gaps in its knowledge \cite{xiaoice}. We found that acknowledging such gaps in our system helped elicit better responses as well, prefacing questions with qualifications like "I can't really tell, but..." or "This might be obvious,  but...". At the intersection of conversational agents and VQA, Visual Dialog introduced a new task, in which machines must hold dialog with a human about visual content \cite{visualdialog}. There was an emphasis in enabling the model to encode conversation history. Our task differs in that we only have one exchange with humans (posing a question), but the authors similarly compiled ground truth (for them, two-person chats) through AMT.
 
 \subsection{State of VQA}
 Finally, we discuss the state of VQA. Two studies examined the deep-learning based approaches for VQA with the best performance and identified shortcomings; they found that today's models are "myopic" (fail on novel instances), jump to conclusions in that they predict an answer before 'listening' to the entire question, and are rigid in that they do not change their answers across images \cite{stateofvqa,vqasurvey}. We hope our work allows VQA systems to become less myopic by seeking out concepts about which they are uncertain.  On the note of systems jumping to conclusions and being rigid, \cite{makingvinvqa} balanced an existing VQA dataset such that every question is associated with a pair of images that are similar but have different answers. They found the top models performed significantly worse on this dataset, meaning they were exploiting language priors \cite{makingvinvqa}. \cite{answertype} attempt to improve performance on VQA by predicting and using answer type, and we take some inspiration by predicting question type to train an attention model that extracts the answer from  a noisy human response.
 %We try to prevent our system from encountering this problem by asking questions that have high entropy among its generated answers, meaning that two similar images can have different answers and we would still pose questions about both and collect them for our dataset. 
 With current VQA datasets being limited \cite{vqasurvey} and performance even then being at around 60 to 70 percent \cite{stateofvqa}, we hope that our work contributes to the expansion of datasets and thus improves performance.

%------------------------------------------------------------------------
\section{Data}

% Describe the data you are working with for your project. What type of data is it? Where did it come from? How much data are you working with? Did you have to do any preprocessing, filtering, or other special treatment to use this data in your project?
We used the Visual Genome dataset~\cite{krishna2017visual}, a computer vision dataset that generalizes ImageNet~\cite{deng2009imagenet} from object categories into relationships between objects. 

We also asked questions to public Instagram users via the comments and collected their responses. Unlike Visual Genome, data gathered from Instagram is much less expensive.

\subsection{Visual Genome}
Visual Genome involved labeling 100,000 images with over two million attributes and two million relationships manually through paid labelers on Amazon Mechanical Turk. The dataset also contains questions and answers regarding various objects and attributes in each image.  We used this part of the dataset to train our VQG and VQA models.

\subsection{Instagram Responses}
Our work focused on parsing human responses to questions we posted on Instagram photos.  We used our VQG and VQA models to ask one question per photo and collected the responses we got.  In total, we collected 120,000 data points of image-question-response sets.  20\% of our data has been labeled by humans with cleaned answers to be used as a test set.

\begin{figure}[h]
    \centering
    \includegraphics[width=0.25\textwidth]{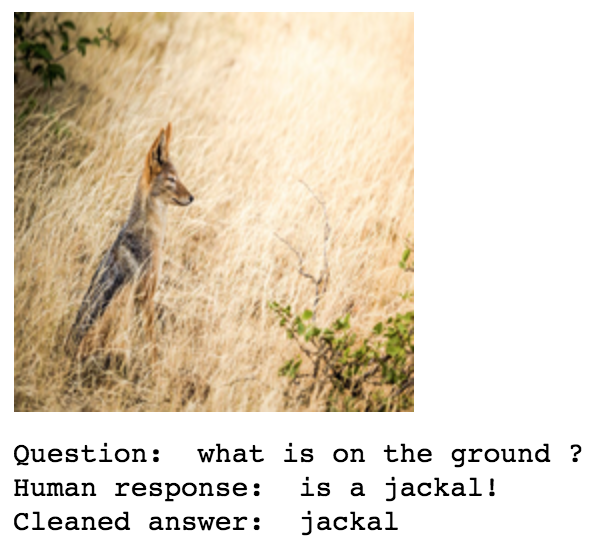}
    \caption{An example data point of Instagram image, question, response, and cleaned answer}
    \label{fig:insta_data}
\end{figure}

\subsection{Text Feature Extraction}
We preprocessed our human Instagram responses to account for common human typing tendencies, such as typos, use of punctuation, and use of emojis. 
We used GloVe word vectors as our main feature representation of our questions and responses ~\cite{pennington2014glove}.  

\begin{figure}[h]
    \centering
    \includegraphics[width=0.25\textwidth]{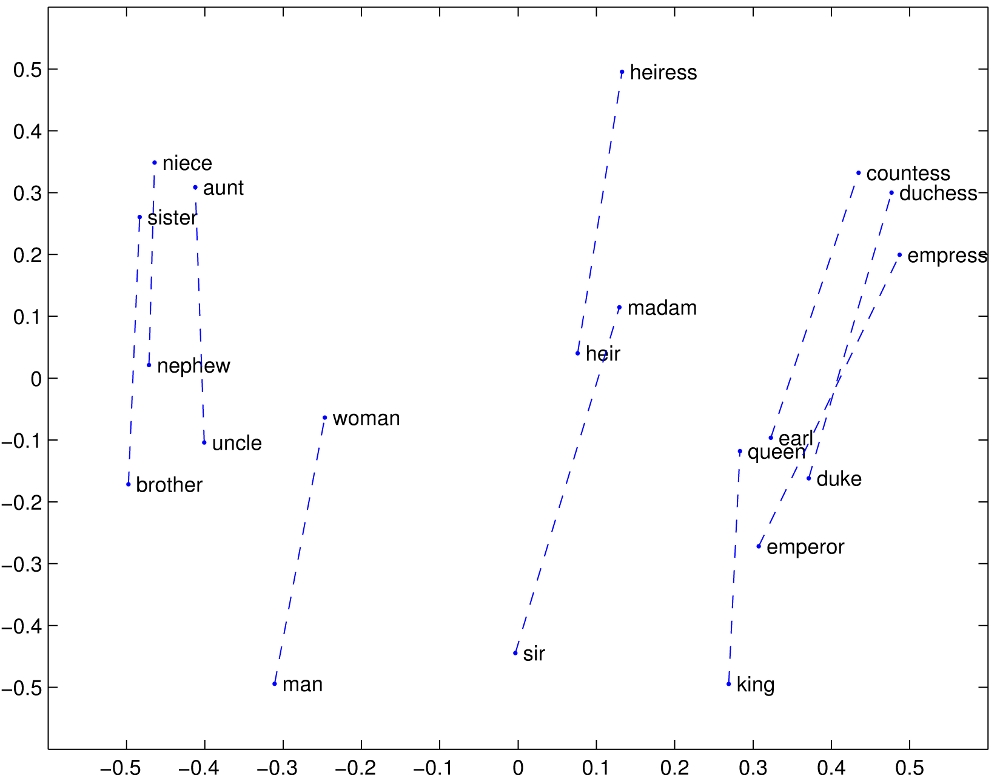}
    \caption{Plotted GloVe vectors in reduced dimensions}
    \label{fig:glove}
\end{figure}

Word vectors are $n$-dimensional representations of words: a numerical way to capture semantic meaning and relationships. In other words, plotting the vectors should show similar words close together.

We not only used GloVe to train word embeddings specific to our tasks, but also used publicly available pre-trained versions (glove.6B specifically).  

\subsection{Image Feature Extraction}
For extracting features from our Visual Genome and Instagram images, we used a ResNet model pre-trained on ImageNet ~\cite{DBLP:journals/corr/HeZRS15}.  We resized our images to $224x224$ and normalized across the data.  

% TODO should we be explaining resnet or imagenet?  where?  how much?

%------------------------------------------------------------------------
\section{Methods}

% Discuss your approach for solving the problems that you set up in the introduction. Why is your approach the right thing to do? Did you consider alternative approaches? You should demonstrate that you have applied ideas and skills built up during the quarter to tackling your problem of choice. It may be helpful to include figures, diagrams, or tables to describe your method or compare it with other methods.

Our main goal was to ask questions and parse clean answers from our noisy human responses on Instagram.  We used a VQG and VQA system for generating questions.  For parsing answers, we propose two different methods: one that focuses on clustering, and one that uses a Recurrent Neural Network.  For each approach, we tried using just the text features as well as combining text features with image features.

\subsection{Baselines}
As baselines, we tried two simple models. 

1. The first extracted key words from the question and the human response based on POS (e.g. removing words from the question like "what is" and keeping "woman wearing") and then calculated the Resnik similarity between each question word and response word. This similarity metric is based on the distance to the least common subsumer of two words in WordNet \cite{wordnet}, and we used it to capture the relationship between what is being asked about and the attribute or predicate that is contained in the answer (e.g. "what animal is in the photo?" would reduce to "animal" and have a Resnik similarity to the word "cat" in the answer that is greater than 0). 

2. The second model was a manual encoding from question type to the POS expected (e.g. "what is the person wearing?" would map to singular or plural noun).

\subsection{Asking Questions with VQG and VQA}

Our approach to generating questions begins by training VQG and VQA models.  These models use ResNets to encode image features, and LSTMs to encode and decode text.  

\begin{figure}[h]
    \centering
    \includegraphics[width=0.4\textwidth]{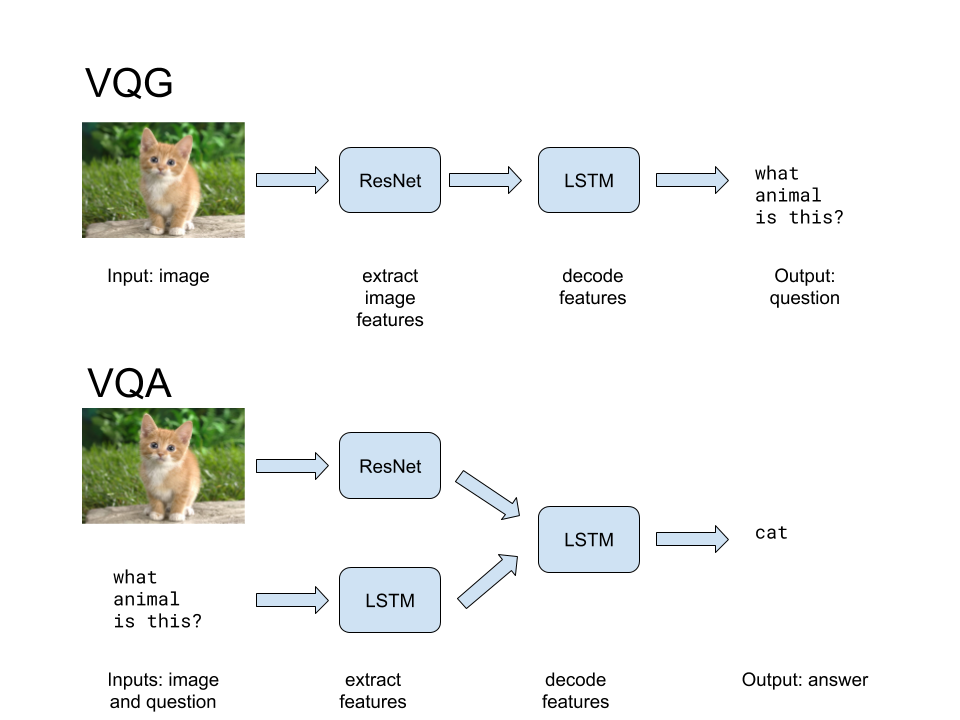}
    \caption{Model diagram for VQG and VQA}
    \label{fig:vqgvqa}
\end{figure}
% https://docs.google.com/drawings/d/1L8tgEXJalj_cDylB8pYwdYNYPMFiEyzBpxYVFeHsES4/edit?usp=sharing

\begin{figure}[h]
    \centering
    \includegraphics[width=0.4\textwidth]{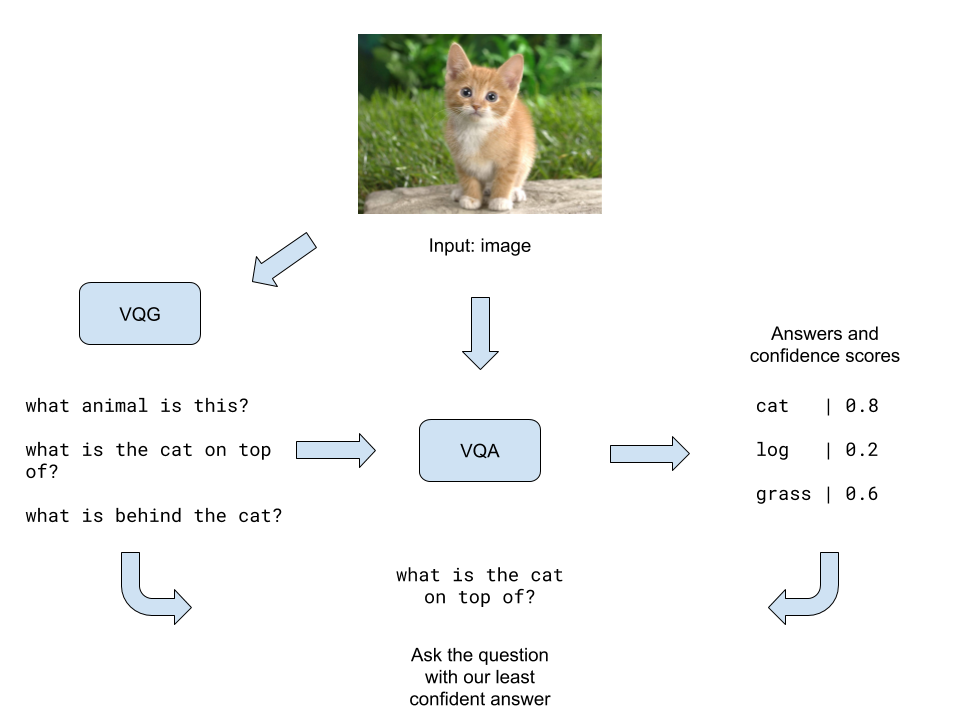}
    \caption{Question Generation and Selection Method}
    \label{fig:q_select}
\end{figure}

As diagrammed in Figure \ref{fig:q_select}, we first generate questions for a given Instagram photo.  Next, we attempt to answer all the questions we generated.  Finally, we select the question with the least confident answer (based on entropy scores) and post that question to the Instagram photo.

% https://docs.google.com/drawings/d/1DsFkZoZNySrZjRcP43tIkGeHlNRpej3EPvj4ZpaNz4Q/edit?usp=sharing

\subsection{Clustering Approach to Parsing}

The first approach begins by clustering responses based on their features from pretrained GloVe embeddings using $K$-Means ~\cite{hartigan+wong:1979}. Our intuition was that similarly noisy human responses would have similar clean answers. We experimented with different values of $K$, and examined the different groups of response embeddings.

\begin{lstlisting}[caption=Location Cluster]
in a parking lot,
in my home,
in chiangmai thailand,
its in esfahan iran,
in my house,
its in ruhengeri,
in my home,
in my clients office somewhere in kl,
in thailand
\end{lstlisting}
We observed that some of the clusters could be semantically interpreted. For example, the above is a sample from a cluster that seemed to be associated with location. We observed that clusters may have latent answer templates; for example, a template for this cluster might be "in \textit{blank}."
    
\begin{figure}[h]
    \centering
    \includegraphics[width=0.5\textwidth]{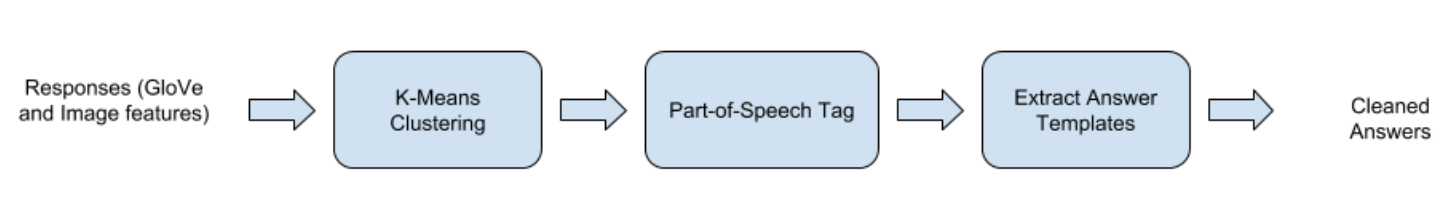}
    \caption{Model diagram for clustering-based answer parsing}
    \label{fig:clustering}
\end{figure}
%https://docs.google.com/drawings/d/1KTQ7rURbB7dOr0VCi98_np0QRbS54Svu3pIyGx7yMlk/edit?usp=sharing

For proof of concept, we used the most commonly occurring POS bigrams and trigrams in the responses as answer templates. For example, these were the templates for the location cluster: 

\begin{lstlisting}[caption=Most common n-grams]
('IN', 'NN')
('IN', 'NN', 'NN')
\end{lstlisting}

"IN" means preposition and "NN" means noun, so the templates were "$<$preposition$>$ $<$noun$>$" or "$<$preposition$>$ $<$noun$>$ $<$noun$>$," like "in vietnam" or "in abruzzo italy."

\subsection{RNN Approach to Parsing}
Rather than clustering responses as in the previous approach, we considered clustering questions. The main insight is that similar questions expect similar answers, so if we train an RNN to input a response and predict question type, we can see where it is attending to in each input to parse the clean answer. In other words, this approach classifies responses by the type of question that yields that response, and our hope is that the attention points to the latent templates. This was a completely unsupervised approach, as our data wasn't labeled with question types. To obtain these types, we again began with clustering, this time on grouping question embeddings and mapping them back to the responses they elicited. This gives us clusters representing question types as well as their counterpart responses.  

\begin{lstlisting}[caption="On" Cluster]
what is on the table ?
what is on the wall ?
what is on the ground ?
what is on top of the pan ?
what is on top of the desserts ?
what is the cat laying on ?
what is on the bench ?
\end{lstlisting}
For example, this cluster of questions seems to capture what was on top of certain items in images.

Then, for each data point, we use the response associated with that question as input to the RNN and the cluster index of the data point as the label. Our task is to predict the type of question that yields each response.

The specific recurrent model we used is the Gated Recurrent Unit (GRU) ~\cite{DBLP:journals/corr/ChoMGBSB14}.  GRUs are considered to be a more generalized version of Long-Short Term Memory (LSTM) cells.  However, we opted for GRUs over LSTMs since GRUs performed better with respect to both accuracy and computation time.

\begin{equation}
\begin{aligned}  z &=\sigma(x_tU^z + s_{t-1} W^z) \\  r &=\sigma(x_t U^r +s_{t-1} W^r) \\  h &= tanh(x_t U^h + (s_{t-1} \circ r) W^h) \\  s_t &= (1 - z) \circ h + z \circ s_{t-1}  
\end{aligned}  
\label{eq:gru}
\end{equation}

GRU cells have two gates: a reset gate and an update gate ($r$ and $z$ in Equation \eqref{eq:gru}).  Intuitively, the reset gate determines how to combine the new input with previous memory, and the reset gate determines how much of the previous memory to keep.  

Once we encode the input using a GRU, we feed the output into an attention module, which can be described as "mapping a query and a set of key-value pairs to an output, where the query, keys, values, output are all vectors" \cite{scaleddpattention}. The output is a weighted sum of the values, and weight is computed by a compatibility function of the query with the corresponding key. The most common types of attention functions are additive and dot-product, but because it can be implemented with matrix multiplication (see equation \eqref{eq:scaleddpattention}, dot-product is much faster and more space-efficient in practice \cite{scaleddpattention}. We used the scaled dot-product attention proposed by \cite{scaleddpattention}, as large values of keys can push the softmax into regions with vanishing gradients. Further, the attention module is a self-attention layer, which has a constant number of sequentially executed operations compared to a recurrent layer which requires \textit{O(n)}.

\begin{equation}
\begin{aligned}  
Attention(Q,K,V) &= softmax(\frac{QK^T}{\sqrt[]{d_k}})V
\end{aligned}  
\label{eq:scaleddpattention}
\end{equation}

Finally, we have a logistic regression classifier on top of the attention outputs to produce our final prediction. 

\begin{figure}[h]
    \centering
    \includegraphics[width=0.27\textwidth]{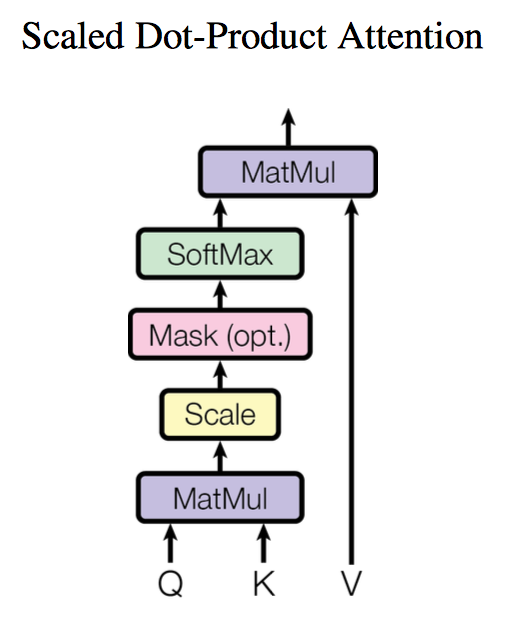}
    \caption{Diagram of the attention module, taken from \cite{scaleddpattention}}
    \label{fig:img_clusters}
\end{figure}

As for how we get the final cleaned answer, we extract all the words that receive at least $\frac{1}{x}$ of the attention that the word with the max attention receives. We experiment with different values of $x$ and discuss the affect of the threshold in the next section.

%------------------------------------------------------------------------
\section{Experiments \& Results}

\subsection{Experiments}

\subsubsection{Instagram Response Rates}
Our main metric for evaluating the usefulness and validity of the questions we post to Instagram is response rate.  We want our questions to be engaging and incite answers.  Asking the questions as our VQG model directly outputted them resulted in an average response rate of 26\%.  

To appear more human-like and be more engaging, we also tried pre-pending questions the questions with exclamations such as "wow!" or "amazing!".  This improved response rates to 31\%.  

\subsubsection{Training Details}
When fitting new GloVe vectors, we capped training iterations at 1000, and their dimensionality at 200.  Our vocabularies had up to 10,000 tokens, so doing much more would take an unreasonable amount of time for marginal improvement. When training the RNN, we split our ground truth data, 70\% for training and 30\% for validation. After fine-tuning, we found that training the model with 1 layer,  300 hidden dimensions, and 140 batch size for 10 epochs at a learning rate of 5e-4 using Adam optimization worked best.

\subsubsection{Evaluation Method}
We compare the answers parsed by our clustering method by calculating the unigram, bigram, trigram, and 4-gram BLEU scores. BLEU (bilingual evaluation understudy) is one of the most popular metrics used in NLP to evaluate the quality of machine-translated text, where quality is measured as "the closer a machine translation is to a professional human translation, the better" \cite{bleu}. In our case, we are translating noisy human responses to clean, pointed answers.

\subsection{Results}

\subsubsection{Baselines}
\begin{table}[h!]
\centering
\caption{Baseline Results}
\label{my-label}
\resizebox{0.5\textwidth}{!}{%
\begin{tabular}{|l|l|l|}
\hline
\textbf{Model} & \textbf{BLEU2} & \textbf{BLEU3} \\ \hline
Baseline 1: WordNet similarity & .0118 & .0126 \\ \hline
Baseline 2: POS expected based on question type & .0131 & .0142 \\ \hline
\end{tabular}%
}
\end{table}
The baselines performed rather poorly, and one explanation may be that they do not consider the order or length of the input when formulating a cleaned answer (e.g. the first simply checks whether a word in the response is similar to a word in the question, and the second only looks at POS). As such, the results are highly prone to noise that decreases the BLEU score.

\subsubsection{Clustering Approach}
\begin{table}[h]
\centering
\caption{BLEU scores for denoising human responses with clustering}
\label{table:clustering}
\resizebox{0.5\textwidth}{!}{%
\begin{tabular}{|l|l|l|}
\hline
\textbf{Number of Clusters} & \textbf{Text Features Only} & \textbf{With Image Features} \\ \hline
K = 4 & 0.4547086627527976 & 0.44208665119310225\\ \hline
K = 5 & 0.470416890081755 & \textbf{0.4484696091199447}\\ \hline
K = 10 & 0.4331685146829659 & 0.4325495610049469 \\ \hline
K = 12 & 0.4888069230753058 & 0.4325495610049469 \\ \hline
K = 15 & \textbf{0.49763641696095107} &  0.42238483996677567 \\ \hline
\end{tabular}%
}
\end{table}

A key observation from the above table is that image features worsened performance overall.  Our overall best performance came from text features only, and text features performed better than including image features for every value of $K$.  Moreover, increasing $K$ decreased performance in the image feature condition.  

\begin{figure}[h]
    \centering
    \includegraphics[width=0.4\textwidth]{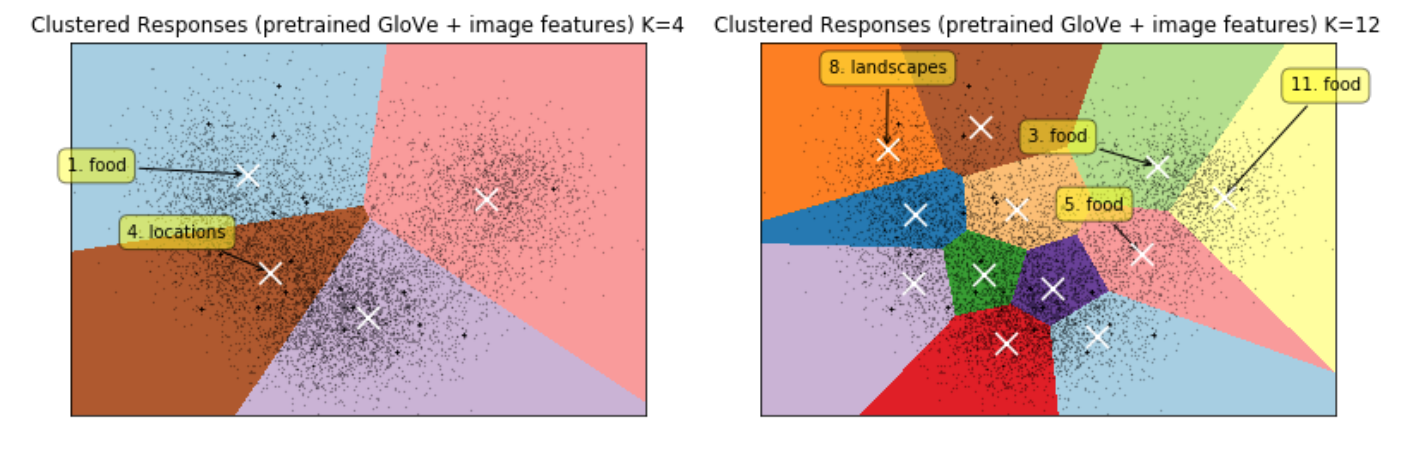}
    \caption{Evolution of PCA-Reduced Response Clusters with Image Features with Increasing K}
    \label{fig:img_clusters}
\end{figure}
We see that with image features included, increasing $K$ shifted the clustering of food-related responses.  We also see multiple food-related clusters, and inconsistent levels of coherence for the other clusters.  

\begin{figure}[h]
    \centering
    \includegraphics[width=0.4\textwidth]{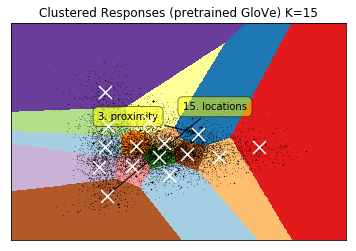}
    \caption{PCA-Reduced Response Clusters for Text Features Only}
    \label{fig:text_cluster}
\end{figure}
On the other hand, in the text features only condition, we see the central clusters divide up the dense pockets of data points quite well.  

Since this model focuses on bi-gram and tri-gram templates, we are only able to extract 2-3 word answers.  

\begin{lstlisting}[caption=Well-parsed answers]
Question:  what is the woman wearing ?
Human response:  she is wearing her work uniform
Correct answer:  ['work', 'uniform']
Model's answer:  ['work', 'uniform']

Question:  where are the tomatoes ?
Human response:  cut in half in the salad
Correct answer:  ['in', 'the', 'salad']
Model's answer:  ['in', 'the', 'salad']
\end{lstlisting}
That said, we parse some of them quite well.  However, longer or shorter answers tend to falter completely, as do answers that use words that appear non-consecutively in the human response words.

\begin{lstlisting}[caption=Failing to parse answers not 2-3 words long]
Question:  what is on the floor ?
Human response:  it is a rug
Correct answer:  ['rug']
Model's answer:  ['it', 'is', 'a']

Question:  what is on top of the plate ?
Human response:  it is a really delicious steak
Correct answer:  ['a', 'steak']
Model's answer:  ['it', 'is', 'a']
\end{lstlisting}

\subsubsection{RNN Attention Approach}

\begin{table}[h]
\centering
\caption{Attention RNN Results}
\label{attn_rnn}
\resizebox{0.5\textwidth}{!}{%
\begin{tabular}{|l|l|l|}
\hline
\textbf{Model} & \textbf{BLEU2} & \textbf{BLEU3} \\ \hline
Attn threshold ratio = 1/2 & 0.5200361205414398 & 0.5535155321168764 \\ \hline
Attn threshold ratio = 1/3 & \textbf{0.540574095667181} & \textbf{0.583688443219591} \\ \hline
\end{tabular}%
}
\end{table}
Overall, we found that using the attention model did improve performance, although we note some observations. First, we found that decreasing the attention threshold ratio improved BLEU score.  However, decreasing the threshold corresponds to being less picky and including more of the response in the cleaned answer.  It follows then that this improves the BLEU scores, because we're simply selecting more words.  Qualitatively, we found the threshold of $1/2$ to produce more pointed answers.  The examples below used a threshold ratio of $1/2$.

\begin{figure}[h]
    \centering
    \includegraphics[width=0.42\textwidth]{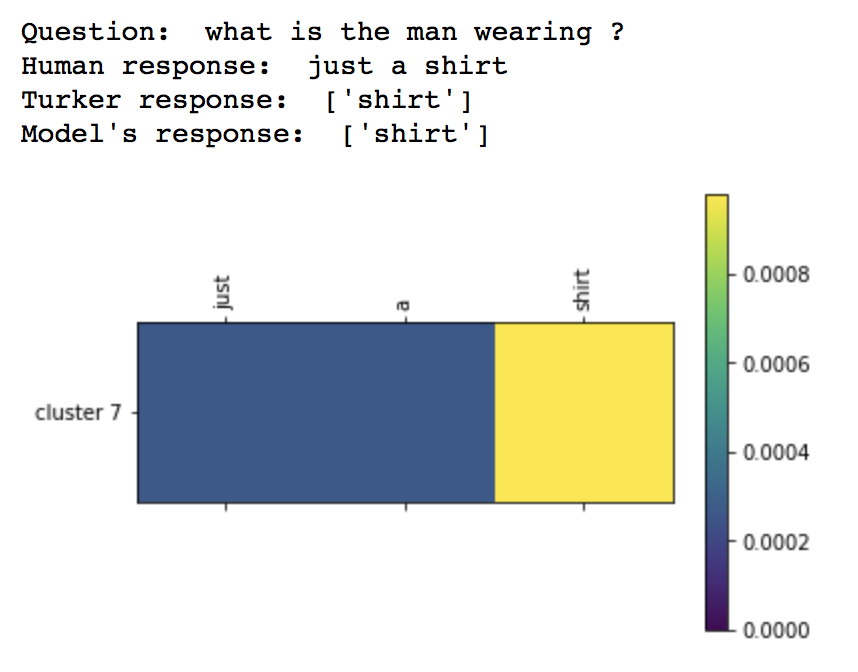}
    \caption{Example 1 from results of attention model. A relatively more simple example but shows how model attends to salient word in the response.}
    \label{fig:shirt_result}
\end{figure}

\begin{figure}[h]
    \centering
    \includegraphics[width=0.45\textwidth]{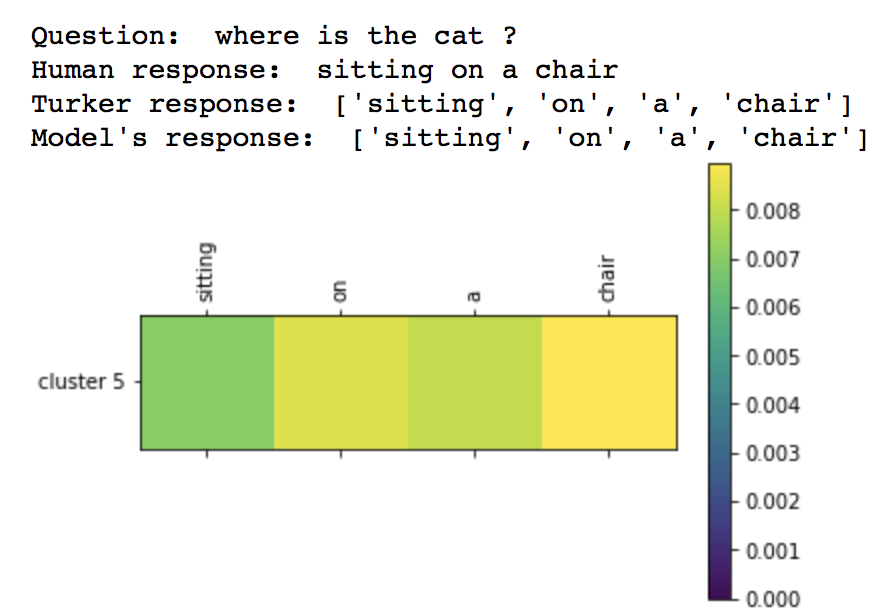}
    \caption{Example 2. An example in which all parts of the answer are relatively salient.}
    \label{fig:cat_result}
\end{figure}

\begin{figure}[h]
    \centering
    \includegraphics[width=0.45\textwidth]{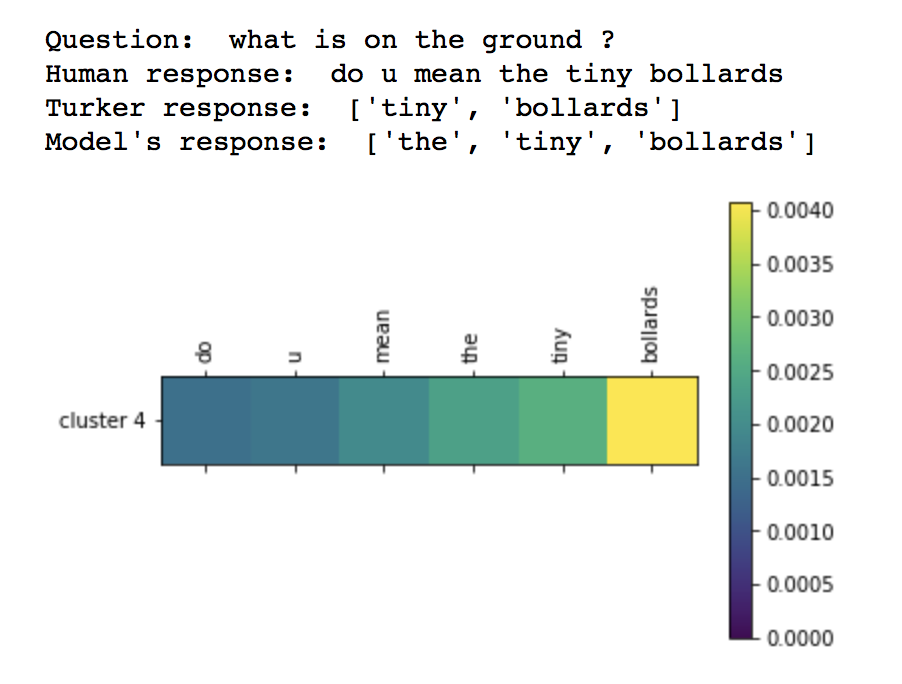}
    \caption{Example 3. Shows how attention model generally leans toward the end of the response.}
    \label{fig:bollards_result}
\end{figure}

Second, there is a general trend of the model attending more to later words in the response. One possible explanation is that many of the human responses begin with "it is the \textit{answer}," "these are all the \textit{answer}," etc. or first mentions the object the question refers to, like "the \textit{object} that you're talking about is \textit{answer}." However, this also means we lose information in longer answers, such as in the example below:

\begin{figure}[H]
    \centering
    \includegraphics[width=0.45\textwidth]{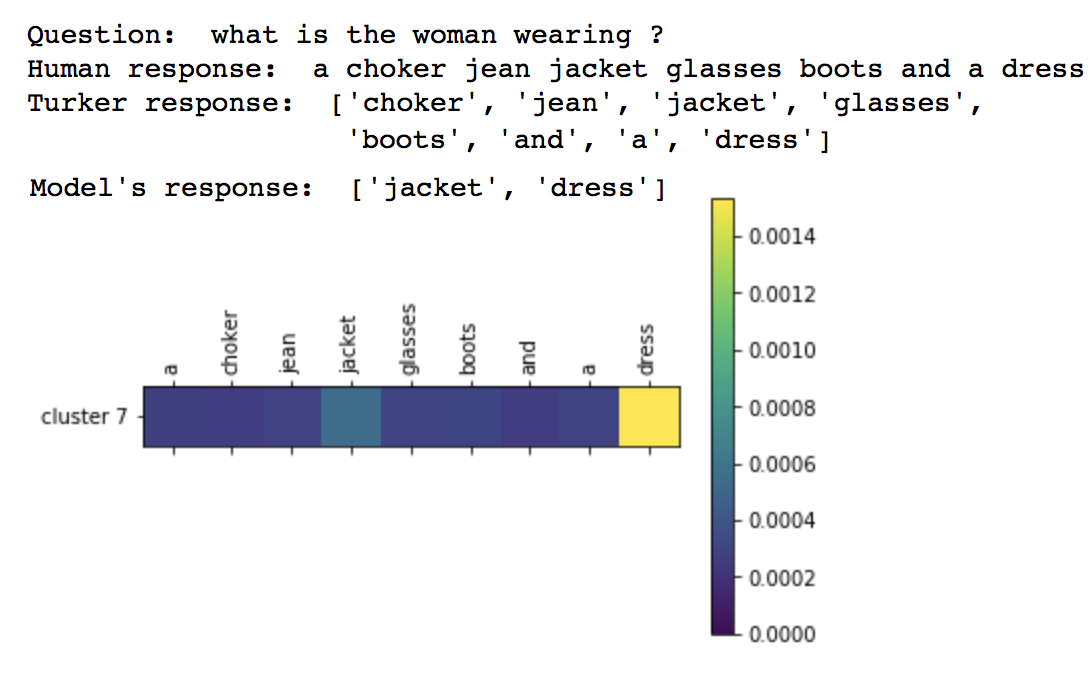}
    \caption{Example 4. Shows how longer answers do not receive as much attention for the earlier words.}
    \label{fig:dress_result}
\end{figure}

Finally, in terms of the model's performance in predicting question type, it achieves 51.84\% accuracy on the validation set, which is much better than random (10\%). Since we have a relatively small training set, we found it was relatively easy to overfit on it, while validation accuracy plateaued; we hope that gathering more training data on the order of hundreds of thousands instead of thousands will help combat this.

%------------------------------------------------------------------------
\section{Conclusion \& Future Work}
% Summarize your report and reiterate key points. Which algorithms were the highest-performing?  Why do you think that some algorithms worked better than others? For future work, if you had more time, more team members, or more compute, what would you explore?

% TODO reiterate results and compare once we have rnn results

Contrary to our initial assumption, adding image features did not improve parsing performance. Our data in general, both the images and the responses, are quite noisy, so it does make sense that adding image features to our embeddings would have translated noise into the clusters and contributed to the decreased performance. Perhaps at greater scale, say on the order of millions instead of ten-thousands, we would be less susceptible to such changes caused by noise. 

As for our attention model, the results illuminated two challenges. The first is that because many of the answers are contained toward the end of the responses, the model becomes biased and gives increasingly more attention as the response progresses, even though the clean answer may begin earlier and/or contain several words. We may overcome this issue by chopping off referential words at the beginning of the response, like "it is the" or "the \textit{object} you're talking about is." Alternatively, we would like to incorporate image features; even though they did not help improve performance in the clustering method, we are interested in testing its effects in the RNN, especially since the prediction accuracy itself could be improved as well. The second challenge is that the BLEU score may not be the best evaluation metric, since simply including more of the words from the original response increases it. As next steps, we will consider other commonly used metrics in NLP, such as METEOR, CIDEr, and ROUGE\_L.

Our main next step is to retrain our VQG/VQA models on our newly cleaned data to try to make use of the acquired knowledge. We still have some challenges,  such as the ground truth data being imperfect. However, this step would close the learning loop and allow the system to continually learn new visual concepts from Instagram and collect new data automatically. We hope that we can achieve this in the future and create a truly ever-expanding dataset.

\section*{Acknowledgments \& Contributions}
%In this section, you must explicitly state what each person on your team did for the project. If you made use of public code (e.g. Github), please provide a link to the original repo. Additionally, you must mention any non-CS 231N collaborators and include a brief sentence on what they did for your project. See the AlphaGo paper’s contributions statement. If you’re part of a research lab and made use of their job scheduling, containerization, or GPUs, briefly include a sentence on this as well.

This project is in conjunction with research advised by Donsuk Lee, Ranjay Krishna, and Professors Michael Bernstein and Fei-Fei Li. Only our own contributions will be submitted for evaluation by the course, and we have received permission and encouragement from our advisers to use our research for the class. Donsuk and Ranjay gave us access to use the dataset of Instagram responses that we acquired together and brainstormed with us when coming up with approaches, as well as providing guidance throughout the research process. The  high-level vision of the project itself is largely the work of Ranjay and our advisers. As for public code, we used:
\begin{itemize}
  \item pretrained GloVe embeddings available \href{https://nlp.stanford.edu/projects/glove/}{here}
  \item scikit-learn's K-Means algorithm found \href{http://scikit-learn.org/stable/modules/generated/sklearn.cluster.KMeans.html}{here}
  \item mttk's public repo for a minimal RNN with self-attention \href{https://github.com/mttk/rnn-classifier}{here}, and 
  \item PyTorch's machine translation tutorial found \href{https://pytorch.org/tutorials/intermediate/seq2seq_translation_tutorial.html}{here}.
\end{itemize}

The authors of this paper, Jihyeon and Sho, worked together on most aspects of the project. Jihyeon implemented the baselines and clustering of responses. Sho implemented the clustering of questions and clustering with image features. Jihyeon implemented the RNN with self-attention and created visualizations of the attentions. Sho created visualizations for the clusters and collected ground truth from the HITs launched on AMT.
%------------------------------------------------------------------------

{\small
\bibliographystyle{ieee}
\bibliography{egbib}
}

\end{document}